%% file: template.tex
\RequirePackage{fix-cm}
\documentclass[smallextended]{llncs}       % onecolumn (second format)
\usepackage{graphicx}
\usepackage{mathptmx}      % use Times fonts if available on your TeX system
\usepackage{breakcites}
\usepackage{amsmath}
\usepackage{amssymb}
\usepackage{floatrow}
\usepackage{subcaption}
\usepackage{paralist}

\newcommand{\E}{\mathbb{E}}
\begin{document}

\title{Guiding InfoGAN with Semi-Supervision}


\author{Adrian Spurr         \and
        Emre Aksan			\and
        Otmar Hilliges
}
\institute{
Advanced Interactive Technologies, ETH Zurich \\
\email{\{adrian.spurr, emre.aksan, otmar.hilliges\}@inf.ethz.ch}
}
\maketitle

%\keywords{Deep generative modeling \and Generative adversarial networks  \and Latent variable models}

%-------------------------------------------------------------------------
\input{sec/abstract}
%-------------------------------------------------------------------------
%-------------------------------------------------------------------------
\input{sec/introduction}
%-------------------------------------------------------------------------
%-------------------------------------------------------------------------
\input{sec/related_work}
%-------------------------------------------------------------------------
%-------------------------------------------------------------------------
\input{sec/method}
%-------------------------------------------------------------------------
%-------------------------------------------------------------------------
\input{sec/implementation}
%-------------------------------------------------------------------------
%-------------------------------------------------------------------------
\input{sec/experiments}
%-------------------------------------------------------------------------
%-------------------------------------------------------------------------
\input{sec/conclusion}
%-------------------------------------------------------------------------
%\begin{acknowledgements}
%\end{acknowledgements}

\bibliographystyle{splncs03}       % APS-like style for physics
\bibliography{references_supInfoGAN}   % name your BibTeX data base

%-------------------------------------------------------------------------
\end{document}

%% file: sec/abstract.tex
\begin{abstract}
In this paper we propose a new semi-supervised GAN architecture (ss-InfoGAN) for image synthesis that leverages information from \emph{few} labels (as little as $0.22\%$, max. $10\%$ of the dataset) to learn semantically meaningful and controllable data representations where latent variables correspond to label categories. The architecture builds on Information Maximizing Generative Adversarial Networks (InfoGAN) and is shown to learn both continuous and categorical codes and achieves higher quality of synthetic samples compared to fully unsupervised settings. Furthermore, we show that using \emph{small} amounts of labeled data speeds-up training convergence. The architecture maintains the ability to disentangle latent variables for which no labels are available.
Finally, we contribute an information-theoretic reasoning on how introducing semi-supervision increases mutual information between synthetic and real data.
\end{abstract}

%% file: sec/introduction.tex
% !TEX root = ../template.tex

\section{Introduction}
\label{intro}
In many machine learning tasks it is assumed that the data originates from a generative process involving complex interaction of multiple independent factors, each accounting for a source of variability in the data.
Generative models are then motivated by the intuition that in order to create realistic data a model must have ``understood'' these underlying factors.
For example, images of handwritten characters are defined by many properties such as character type, orientation, width, curvature and so forth.

Recent models that attempt to extract these factors are either completely supervised \cite{salimans2016improved,odena2016,mirza2014} or entirely unsupervised \cite{desjardins2012, chen2016}. Supervised approaches allow for extraction of the desired parameters but require fully labeled datasets and a priori knowledge about which factors underlie the data. However, factors not corresponding to labels will not be discovered.
In contrast, unsupervised approaches require neither labels nor a priori knowledge about the the underlying factors but this flexibility comes at a cost: such models provide no means of exerting control on what kind of features are found. For example, Information Maximizing Generative Adversarial Networks (InfoGAN) have recently been shown to learn disentangled data representations. Yet the extracted representations are not always directly interpretable by humans and lack direct measures of control due to the unsupervised training scheme. Many application scenarios however require control over \emph{specific} features.

Embracing this challenge, we present a new semi-supervised generative architecture that requires only few labels to provide control over which factors are identified. Our approach can exploit already existing labels or use datasets that are augmented with easily collectible labels (but are not fully labeled). The model, based on the related InfoGAN \cite{chen2016} is dubbed semi-supervised InfoGAN (ss-InfoGAN).
In our approach we maximize two mutual information terms: \begin{inparaenum}[i)]
\item The mutual information between a code vector and real labeled samples, guiding the corresponding codes to represent the information contained in the labeling,
\item and the mutual information between the code vector and the synthetic samples.
\end{inparaenum}
By doing so ss-InfoGAN can find representations that unsupervised methods such as InfoGAN fail to find, for example the category of digits of the SVHN dataset.
Notably our approach requires only $10\%$ of labeled data for the hardest dataset we tested and for simpler datasets  only $132$ labeled samples ($0.22\%$) were necessary.

We discuss our method in full, provide an information theoretical rationale for the chosen architecture and demonstrate its utility in a number of experiments on the MNIST \cite{lecun1998}, SVHN \cite{netzer2011}, CelebA \cite{liu2015} and CIFAR-10 \cite{krizhevsky2009} datasets. We show that our method improves results over the state-of-the-art, combining advantages of supervised and unsupervised approaches.

%% file: sec/related_work.tex
% !TEX root = ../template.tex

\section{Related Work}
\label{sec:related_work}
Many approaches to modeling the data generating process and identifying the underlying factors by learning to synthesize samples from disentangled representations exist. An example of an early approach is supervised bi-linear models \cite{tenenbaum2000}, separating style from the content. Zhu et al. \cite{zhu2014} use a multi-view deep perceptron model to untangle the identity and viewpoint of face images. Weakly supervised methods based on supervised clustering, have been proposed such as high-order Boltzman machines \cite{reed2014} applied on face images.

Variational Autoencoders (VAEs) \cite{kingma2013auto} and Generative Adversarial Networks (GANs) \cite{goodfellow2014} have recently seen a lot of interest in generative modeling problems. In both approaches a deep neural network is trained as a generative model by using standard backpropagation, enabling synthesis of novel samples without explicitly learning the underlying data distribution. VAEs maximize a lower bound on the marginal likelihood which is expected to be tight for accurate modeling \cite{kingma2016improving,burda2015importance,sonderby2016ladder,siddarth2017}.
In contrast, GANs optimize a minimax game objective via a discriminative adversary. However, they have been shown to be unstable and fragile \cite{salimans2016improved, metz2016unrolled}.

Employing semi-supervised learning, Kingma et al. \cite{kingma2014} use VAEs to isolate content from other variations, and achieve competitive recognition performance in addition to high-quality synthetic samples. Deep Convolutional Inverse Graphics Network (DC-IGN) \cite{kulkarni2015}, which uses a VAE architecture and a specially tailored training scheme is capable of learning a disentangled latent space in fully supervised manner. Since the model is evaluated by using images of 3D models, labels for the underlying factors are cheap to attain. However, this type of dense supervision is unfeasible for most non-synthetic datasets.

Adversarial Autoencoders \cite{makhzani2015} combine the VAE and GAN frameworks in using an adversarial loss on the latent space. Similarly, Mathieu et al. \cite{mathieu2016disentangling} introduces an adversarial loss on the reconstructions of VAE, that is, on the pixel space. Both models are shown to learn both discrete and continuous latent representations and to disentangle style and content in images. However, these hybrid architectures have conceptually different designs as opposed to GANs. While the former learns the data distribution via Autoencoder training and employ the adversarial loss as a regularizer, the latter directly relies on an adversarial objective. Despite the robust and stable training, VAEs have tendency to generate blurry images \cite{lamb2016discriminative}.

Conditional GANs \cite{salimans2016improved,odena2016,mirza2014} augment the GAN framework by using class labels. Mirza and Osindero \cite{mirza2014} train a class-conditional discriminator while \cite{salimans2016improved} and \cite{odena2016} use auxiliary loss terms for the labels. Salimans et al. \cite{salimans2016improved} use conditional GANs for pre-training, aiming to improve semi-supervised classification accuracy of the discriminator. Similarly, the AC-GAN model \cite{odena2016} introduces an additional classification task in the discriminator to provide class-conditional training and inference of the generator in order to be able to synthesize higher resolution images than previous architectures.
Our work is similar to the above in that it provides class-conditional generation of images. However, due to MI loss terms our architecture can
\begin{inparaenum}[i)]
\item be employed in both supervised \emph{and} semi-supervised settings.
\item can learn interpretable representations in addition to smooth manifolds and
\item can exploit continuous supervision signals if such labels are available.
\end{inparaenum}

Comparatively fewer works treat the subject of fully unsupervised generative models to retrieve interpretable latent representations. Desjardins et al. \cite{desjardins2012} introduced a higher-order RBM for recognition of facial expressions. However, it can only disentangle discrete latent factors and the computational complexity rises exponentially in the number of features. More recently, Chen et al. \cite{chen2016} developed an extension to GANs, called Information Maximizing Generative Adversarial Networks (InfoGAN). It enforces the generator to learn disentangled representations through increasing the mutual information between the synthetic samples and a newly introduced latent code. Our work extends InfoGAN such that additional information can be used. Supervision can be a necessity if the model struggles in learning desirable representations or if \emph{specific} features need to be controlled by the user. Our model provides a framework for semi-supervision in InfoGANs. We find that leveraging few labeled samples brings improvements on the convergence rate, quality of representations and synthetic samples. Moreover, semi-supervision helps the model in capturing otherwise difficult to capture representations.

%% file: sec/method.tex
% !TEX root = ../template.tex

\section{Method}
\subsection{Preliminaries: GAN and InfoGAN}
In the GAN framework, a generator $G$ producing synthetic samples is pitted against a discriminator $D$ that attempts to discriminate between real data and samples created by $G$. The goal of the generator is to match the distribution of generated samples $P_G$ with the real distribution $P_{data}$. Instead of explicitly estimating $P_G(x)$, $G$ learns to transform noise variables $z \sim P_{noise}$ into synthetic samples $\tilde{x} \sim P_G$.
The discriminator $D$ outputs a single scalar $D(x)$ representing the probability of a sample $x$ coming from the true data distribution.
Both $G(z;\theta_g)$ and $D(x;\theta_d)$ are differentiable functions parametrized by neural networks. We typically omit the parameters $\theta_g$ and $\theta_d$ for brevity.
$G$ and $D$ are simultaneously trained by using the minimax game objective $V_{GAN}(D,G)$:
\begin{align}
\begin{split}
\label{eq:GAN}
\min_G\max_D V_{GAN}(D,G) = \E_{x\sim P_{data}}[\log D(x)] + \E_{z\sim P_{noise}}[\log (1-D(G(z)))]
\end{split}
\end{align}

GANs map from the noise space to data space without imposing any restrictions. This allows $G$ to produce arbitrary mappings and to learn highly dependent factors that are hard to interpret. Therefore, variations of $z$ in any dimension often yields entangled effects on the synthetic samples $\tilde{x}$. InfoGANs \cite{chen2016} are capable of learning disentangled representations. InfoGAN extends the unstructured noise $z$ by introducing a latent code $c$. While $z$ represents the incompressible noise, $c$ describes semantic features of the data. In order to prevent $G$ from ignoring the latent codes $c$, InfoGAN regularizes learning via an additional cost term penalizing low mutual information between $c$ and $\tilde{x} = G(z,c)$:
\begin{align}
\begin{split}
\label{eq:objFun_infoGAN}
\min_{G,Q}\max_D V_{InfoGAN}(D,G,Q,\lambda_1) &= V_{GAN}(D,G) - \lambda_1 L_I(G,Q), \\
I(C;\tilde{X}) \geq L_I(G,Q) &= \E_{c \sim P_{c}, \tilde{x}\sim P_G}[\log Q(c|\tilde{x})] + H(c),
\end{split}
\end{align}
where $Q$ is an auxiliary parametric distribution approximating the posterior $P(c|x)$, $L_I$ corresponds to the lower bound of the mutual information $I(C;\tilde{X})$ and $\lambda_1$ is the weighting coefficient.

\subsection{Semi-Supervised InfoGAN}
\begin{figure}[h]
\centering
	  \includegraphics[width=0.6\linewidth]{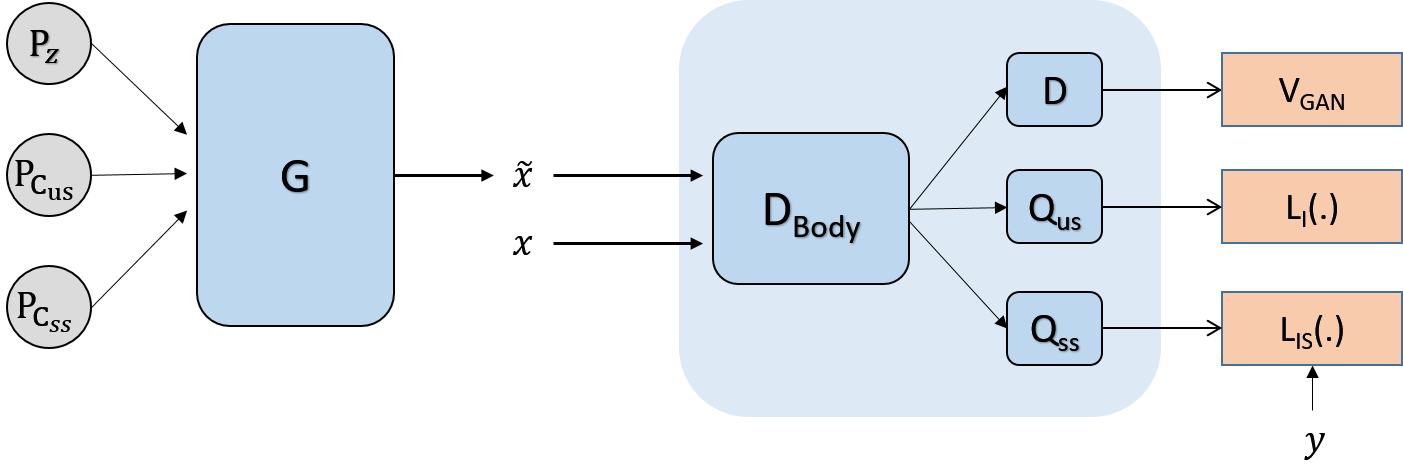}
	  \caption{Schematic overview of the ss-InfoGAN network architecture. $P_z$, $P_{C_{us}}$ and $P_{C_{ss}}$ are the distributions of the noise and latent variables $z$, $c_{us}$ and $c_{ss}$, respectively.}
	  \label{fig:ssinfogan}
\end{figure}
Although InfoGAN can learn to disentangle representations in an unsupervised manner for simple datasets, it struggles to do so on more complicated datasets such as CelebA or CIFAR-10. In particular, capturing categorical codes is challenging and hence InfoGAN yields poorer performance in class-conditional generation task than competing methods. Moreover, depending on the initialization, the learned latent codes may differ between training sessions further reducing interpretability.

In Semi-Supervised InfoGAN (ss-InfoGAN) we introduce available or easily acquired labels to address these issues. Figure \ref{fig:ssinfogan} schematically illustrates our architecture.
To make use of label information we decompose the latent code $c$ into a set of semi-supervised codes, $c_{ss}$, and unsupervised codes, $c_{us}$, where $c_{ss} \cup c_{us} = c$.
The semi-supervised codes encode the same information as the labels $y$, whereas $c_{us}$ are free to encode potential remaining semantic factors.

We seek to increase the mutual information $I(C_{ss};X)$ between the latent codes $c_{ss}$ and the labeled real samples $x$, by interpreting labels $y$ as the latent codes $c_{ss}$, (i.e. $y=c_{ss}$). Note that not all samples need to be labeled for the generator to learn the inherent semantic meaning of $y$. 
We additionally want to increase the mutual information $I(C_{ss};\tilde{X})$ between the semi-supervised latent codes and the synthetic samples $\tilde{x}$ so that information can flow back to the generator.
This is accomplished via Variational Information Maximization \cite{barber2003algorithm} in deriving lower bounds for both MI terms. For the lower bounds of $I(C_{ss};\cdot)$ we utilize the same derivation as InfoGAN:
\begin{align}
\begin{split}
I(C_{ss};X) &\geq \E_{c \sim P_{C_{ss}}, x\sim P_X}[\log Q_1(c_{ss}|x)] + H(C_{ss}) = L^1_{IS}(Q_1),
\end{split}
\\
\begin{split}
I(C_{ss};\tilde{X}) &\geq \E_{c \sim P_{C_{ss}}, \tilde{x}\sim P_G}[\log Q_2(c_{ss}|\tilde{x})] + H(C_{ss}) = L^2_{IS}(Q_2, G),
\end{split}
\end{align}
where $Q_1$ and $Q_2$ are again auxiliary distributions to approximate posteriors and are parametrized by neural networks. With $Q_1 = Q_2 = Q_{ss}$ we attain the MI cost term:
\begin{align}
L_{IS}(Q_{ss}, G) = L^1_{IS}(Q_{ss}) + L^2_{IS}(Q_{ss}, G)
\end{align}

Since we would like to encode the labels $y$ via latent codes $c_{ss}$, we optimize $L^1_{IS}(Q_{ss})$ with respect to $Q_{ss}$ and $L^2_{IS}(Q_{ss}, G)$ only with respect to $G$. The final objective function is then:
\begin{align}
\label{eq:objFun_SSinfoGAN}
\min_{G,Q_{us},Q_{ss}}\max_D \quad &V_{ss\textsc{-}InfoGAN}(D,G,Q_{us},Q_{ss},\lambda_1,\lambda_2) \\ = &V_{InfoGAN}(D,G,Q_{us},\lambda_1) - \lambda_2 L_{IS}(G, Q_{ss})
\end{align}

Training $Q_{ss}$ on labeled real data $(x, y)$ enables $Q_{ss}$ to encode the semantic meaning of $y$ via $c_{ss}$ by means of increasing the mutual information $I(C_{ss};X)$.
Simultaneously, the generator $G$ acquires the information of $y$ indirectly by increasing $I(C_{ss};\tilde{X})$ and learns to utilize the semi-supervised representations in synthetic samples.
In our experiments we find that a small subset of labeled samples is enough to observe significant effects.

We show that our approach gives control over discovered properties and factors and that our method achieves better image quality. Here we provide an information theoretic underpinning shedding light on the reason for these gains.
By increasing both $I(C_{ss};X)$ and $I(C_{ss};\tilde{X})$, the mutual information term $I(X;\tilde{X})$ is increased as well. We make the following assumptions:
\begin{align}
\label{eq:assumptions1}
X \leftarrow C_{ss} \rightarrow \tilde{X}, \\
\label{eq:assumptions2}
I(X;\tilde{X}) = 0 \text{ initially}, \\
\label{eq:assumptions3}
H(C_{ss}) = \mathtt{C},
\end{align}
where $\mathtt{C}$ is a constant and $\rightarrow$ are dependency relations. Assumption (\ref{eq:assumptions1}) follows the intuition that the data is hypothesized to arise from the interaction of independent factors. While latent factors consist of $z$, $C_{us}$ and $C_{ss}$, we abstract for the sake of simplicity. Assumption (\ref{eq:assumptions2}) formulates the initial state of our model where the synthetic data distribution $P_G$ and the data distribution $P_{data}$ are independent.
Finally we can assume that labels follow a fixed distribution and hence have a fixed entropy $H(C_{ss})$, giving rise to (\ref{eq:assumptions3}).

We decompose $H(C_{ss})$ and reformulate $I(X;\tilde{X})$ in the following way:
\begin{align}
\label{eq:entropy_decomposed}
&H(C_{ss}) = I(C_{ss};X) + I(C_{ss};\tilde{X}) + H(C_{ss}|X,\tilde{X}) - I(C_{ss};X;\tilde{X}), \\
\label{eq:multivariate_mi}
&I(C_{ss};X;\tilde{X}) =  I(X;\tilde{X}) - I(X;\tilde{X}|C_{ss})
\end{align}
where $I(C_{ss};X;\tilde{X})$ is the multivariate mutual information term.
While pointwise MI is per definition non-negative, in the multivariate case negative values are possible if two variables are coupled via the third.
By using the conditional independence assumption (\ref{eq:assumptions1}), we have
\begin{equation}
\label{eq:mutual_information_ge0}
I(C_{ss};X;\tilde{X}) =  I(X;\tilde{X}) - I(X;\tilde{X}|C_{ss}) = I(X;\tilde{X}) \ge 0.
\end{equation}
Thus the entropy term $H(C_{ss})$ in Eq. (\ref{eq:entropy_decomposed}) takes the form
\begin{equation}
\begin{split}
H(C_{ss}) &= I(C_{ss};X) + I(C_{ss};\tilde{X}) + H(C_{ss}|X,\tilde{X}) - I(X;\tilde{X})
\end{split}
\end{equation}
Let $\Delta$ symbolize the change in value of a term. According to assumption (\ref{eq:assumptions3}), the following must hold:
\begin{equation}
\Delta_{I(C_{ss};X)} + \Delta_{I(C_{ss};\tilde{X})} + \Delta_{H(C_{ss}|X,\tilde{X})} - \Delta_{I(X;\tilde{X})} = 0
\end{equation}
Note that $\Delta_{I(C_{ss};X)}$ and $\Delta_{I(C_{ss};\tilde{X})}$ increase during training since we directly optimize these terms, leading to the following cases:
\begin{equation}
\begin{split}
\Delta_{I(C_{ss};X)} + \Delta_{I(C_{ss};\tilde{X})} \geq -\Delta_{H(C_{ss}|X,\tilde{X})} \implies& \Delta_{I(X;\tilde{X})} \geq 0 \\
\Delta_{I(C_{ss};X)} + \Delta_{I(C_{ss};\tilde{X})} < -\Delta_{H(C_{ss}|X,\tilde{X})} \implies& \Delta_{I(X;\tilde{X})} < 0
\end{split}
\end{equation}
The first case results in the desired behavior. However the latter case cannot occur, as it would result in negative mutual information $I(X;\tilde{X})$. Hence, based on our assumptions, increasing both $I(C_{ss};X)$ and $I(C_{ss};\tilde{X})$ leads to an increase in $I(X;\tilde{X})$.

%% file: sec/implementation.tex
% !TEX root = ../template.tex

\section{Implementation}
For both $D$ and $G$ of ss-InfoGAN we use a similar architecture with DCGAN \cite{radford2015}, which is reported to stabilize training. The networks for the parametric distributions $Q_{us}$ and $Q_{ss}$ share all the layers with $D$ except the last layers. This is similar to \cite{chen2016}, which models $Q$ as an extension to $D$. This approach has the disadvantage of negligibly higher computational cost for $Q_{ss}$ in comparison to InfoGAN. However, this is offset by a faster convergence rate in return.

In our experiments with low amount of labeled data, we initially favor drawing labeled samples, which improves convergence rate of the supervised latent codes significantly. During training the probability of drawing a labeled sample is annealed until the actual labeled sample ratio in the data is reached. The loss function used to calculate $L_I$ and $L_{IS}$ is the cross-entropy for categorical latent codes and the mean squared error for continuous latent codes.
The unsupervised categorical codes are sampled from a uniform categorical distribution whereas the continuous codes are sampled from a uniform distribution.
All the experimental details are listed in the supplementary document. In the interest of reproducible research, we provide the source code on GitHub. \footnote{Implementation can be found at https://github.com/spurra/ss-infogan}

For comparison we re-implement the original InfoGAN architecture in the Torch framework \cite{collobert2011} with minor modifications. Note that there may be differences in results due to the unstable nature of GANs, possibly amplified by using a different framework and different initial conditions. In our implementation the loss function for continuous latent codes are not treated as a factored Gaussian, but approximated with the mean squared error, which leads to a slight adjustment in the architecture of $Q$.

%% file: sec/experiments.tex
% !TEX root = ../template.tex
\section{Experiments}
In our study we focus on interpretability of the representations and quality of synthetic images under different amount of labeling. The closest related work to that of ours is InfoGAN, and the aim was to directly improve upon that architecture. The existing semi-supervised generative modeling studies on the other hand, aim to learn discriminative representations for classification. Therefore we make a direct comparison with InfoGAN.

We evaluate our model on the MNIST \cite{lecun1998}, SVHN \cite{netzer2011}, CelebA \cite{liu2015} and CIFAR-10 \cite{krizhevsky2009} datasets. First, we inspect how well ss-InfoGAN learns the representations as defined by existing labels. Second, we qualitatively evaluate the representations learned by ss-InfoGAN. Finally, we analyze how much labeled data is required for the model to encode semantic meaning of the labels $y$ via $c_{ss}$.

We hypothesize that the quality of the generator in class-conditional sample synthesis can be quantitatively assessed by a separate classifier trained to recognize class labels. The class labels of the synthetic samples (i.e. the class conditional inputs of the generator) are regarded as true targets and compared with the classifier's predictions. In order to prevent biased results due to the generator overfitting, we train the classifier $C$ by using the \textit{test set}, and validate on the \textit{training set} for each dataset. Despite the \textit{test set} consisting fewer samples, the classifier $C$ generally performs well on the unseen \textit{training set}. In our experiments, we use a standard CNN (architecture described in the supplementary file) for the MNIST, CelebA and SVHN datasets and Wide Residual Networks \cite{zagoruyko2016} for CIFAR-10 dataset.

In order to evaluate how well the model separates types of semantic variation, we generate synthetic images by varying only one latent factor by means of linear interpolation while keeping the remaining latent codes fixed.

To evaluate the necessary amount of supervision we perform quantitative analysis of the classifier accuracy and qualitative analysis by examining synthetic samples. To do so, we discard increasingly bigger sets of labels from the data. Note that $Q_{ss}$ is trained only by using labeled samples and hence sees less data, whereas the rest of the architecture, namely the generator and the shared layers of the discriminator, uses the entire training samples in unsupervised manner. The minimum amount of labeled samples required to learn the representation of labels $y$ varies depending on the dataset. However, for all our experiments it never exceeded $10\%$.

\begin{figure}[!htb]
\begin{subfigure}{.44\textwidth}
\centering
	  \includegraphics[width=0.9\linewidth]{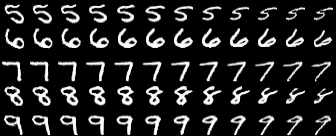}
 	 \caption{$0.22\%$, orientation}
     \label{fig:mnist_1a}
\end{subfigure}%
\begin{subfigure}{.44\textwidth}
\centering
	  \includegraphics[width=0.9\linewidth]{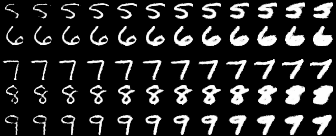}
 	 \caption{$0.22\%$, width}
     \label{fig:mnist_1b}
\end{subfigure}

\begin{subfigure}{.44\textwidth}
\centering
	  \includegraphics[width=0.9\linewidth]{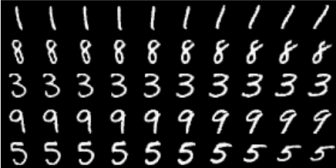}
 	 \caption{$0\%$, orientation}
     \label{fig:mnist_1c}
\end{subfigure}
\begin{subfigure}{.44\textwidth}
\centering
	  \includegraphics[width=0.9\linewidth]{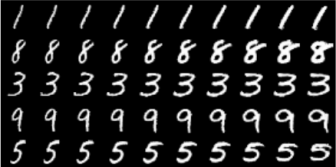}
 	 \caption{$0\%$, width}
     \label{fig:mnist_1d}
\end{subfigure}%
  	\caption{\textbf{Manipulating latent code on MNIST:} \textit{In all figures of latent code manipulation we use the convention that a latent code varies from left to right ($x$-axis) while the remaining codes and the noise are kept fixed. Each row along the $y$-axis corresponds to a categorical latent code encoding a class label unless otherwise stated. The interpretation of the varying latent code is provided under the image.} Synthetic images generated by interpolating the latent codes, encoding the digit "orientation" and "width", between -2 and 2. (\ref{fig:mnist_1a}, \ref{fig:mnist_1b}) ss-InfoGAN with $0.22\%$ supervision. (\ref{fig:mnist_1c}, \ref{fig:mnist_1d}) InfoGAN (taken from \cite{chen2016}).}
	\label{fig:mnist_1}
\end{figure}
%%%%%%%%%%%%%% MNIST
\subsection{MNIST}

\begin{figure}[!htb]
  % Equal length
  \hspace*{\fill}%
  \subcaptionbox{MNIST\label{fig:mnist_2}}{\includegraphics[width=6cm]{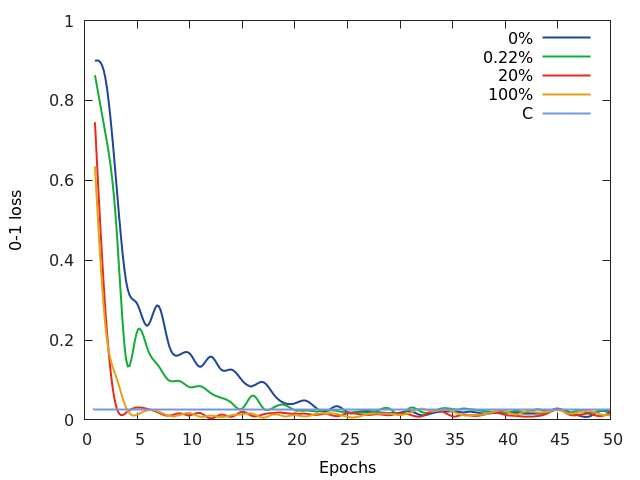}}\hfill%
  \subcaptionbox{SVHN\label{fig:svhn_3}}{\includegraphics[width=6cm]{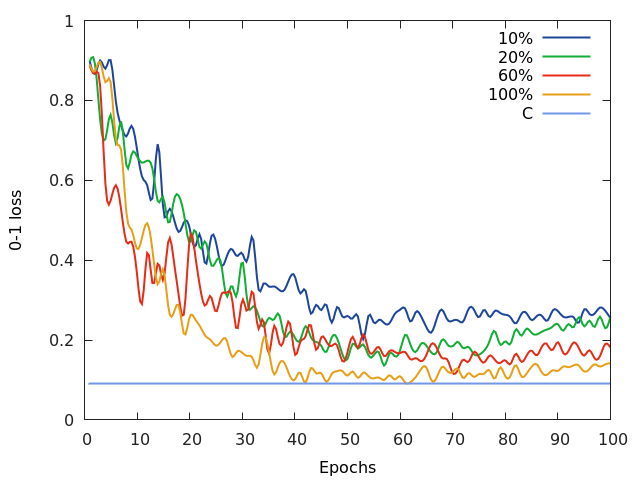}}%
  \hspace*{\fill}%
  \caption{\textbf{0-1 loss on synthetic samples:} \textit{In all 0-1 loss figures we plot the classification accuracy on synthetic samples of the respective dataset. During training of ss-InfoGAN, a batch of synthetic samples are randomly generated, and evaluated by the independent classifier $C$. Colors represent the GAN models trained with different amount of supervision and classifier performance (C) on real validation samples.}}
\end{figure}

MNIST is a standard dataset used to evaluate many generative models. It consists of handwritten digits, and is labeled with the digit category. Figure \ref{fig:mnist_1} presents the synthetic samples generated with our model and InfoGAN by varying the latent code. Due to lower complexity of the dataset, InfoGAN is capable of learning the digit representation unsupervised. However, using just $0.22\%$ of the available data has a two-fold benefit. First, semi-supervision provides additional fine-grained control (e.g., digits are already sorted in ascending manner in Figure \ref{fig:mnist_1a}, \ref{fig:mnist_1b}). Second, we experimentally verified that the additional information increases convergence speed of the generator, illustrated in Figure \ref{fig:mnist_2}. The 0-1 loss of the classifier $C$ decreases faster as more labeled samples are introduced while the fully unsupervised setting (i.e. InfoGAN) is the slowest. The smallest amount of labeled samples for which the effect of supervision is observable is $0.22\%$ of the dataset, which corresponds to $132$ labeled samples out of $60'000$.
%%%%%%%%%%%%%%% SVHN
\subsection{SVHN}
\begin{figure}[!htb]
  % Equal length
  \subcaptionbox{$10\%$, red gradient\label{fig:svhn_1a}}{\includegraphics[width=4.4cm]{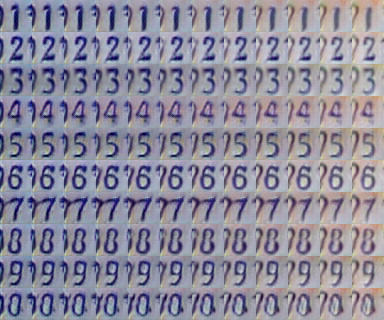}}\hspace{1em}%
  \subcaptionbox{$10\%$, brightness\label{fig:svhn_1b}}{\includegraphics[width=4.4cm]{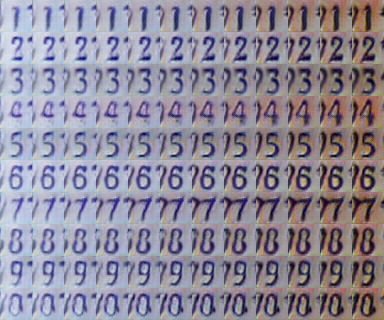}}
  \subcaptionbox{$100\%$, brightness\label{fig:svhn_1c}}{\includegraphics[width=4.4cm]{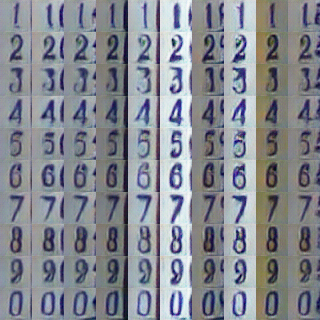}}\hspace{1em}%
  \subcaptionbox{$100\%$, digit font\label{fig:svhn_1d}}{\includegraphics[width=4.4cm]{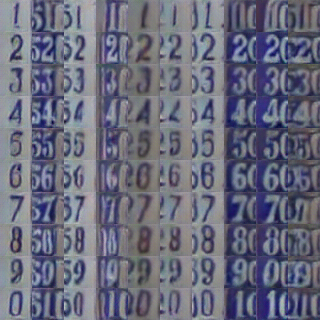}}%
  \caption{\textbf{Manipulating latent code on SVHN:} Latent codes encoding the "brightness", "digit font" and "red gradient" are interpolated between -2 and 2 for each semi-supervised categorical code. (\ref{fig:svhn_1a}, \ref{fig:svhn_1b}) ss-infoGAN with $10\%$ supervision. (\ref{fig:svhn_1c}, \ref{fig:svhn_1d})  ss-InfoGAN with $100\%$ supervision.}
	\label{fig:svhn_1}
\end{figure}
\begin{figure}[!htb]
  % Equal length
  \hspace*{\fill}%
  \subcaptionbox{Real samples}{\includegraphics[height=4.6cm]{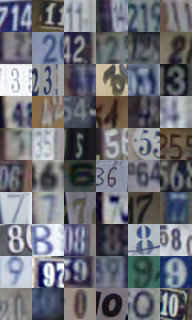}}\hfill%
  \subcaptionbox{$0\%$ \label{svhn_2b}}{\includegraphics[height=4.6cm]{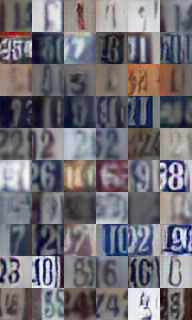}}\hfill%
  \subcaptionbox{$10\%$}{\includegraphics[height=4.6cm]{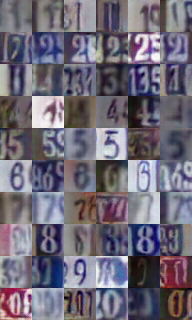}}\hfill%
  \subcaptionbox{$100\%$}{\includegraphics[height=4.6cm]{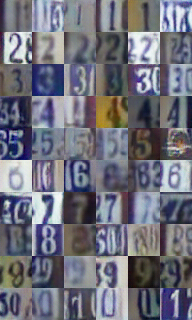}}%
  \hspace*{\fill}%
  \caption{\textbf{Random synthetic SVHN samples:} \textit{In all figures of randomly synthesized images we present examples of real samples from the dataset and synthetic images. Models are trained with different amount of supervision which is noted under the images, where the $0\%$ supervision corresponds to InfoGAN.} In each row, the semi-supervised categorical code encoding the digits is kept fixed while rest of the input vector, (i.e. $z$, $c_{us}$) and the remaining codes in $c_{ss}$, is randomly drawn from the latent distribution. Although each row represents a digit the fully unsupervised model (\ref{svhn_2b}) (i.e. InfoGAN) lacks control on the digit category.}
	\label{fig:svhn_2}
\end{figure}
Next, we run ss-InfoGAN on the SVHN dataset which consists of color images, hence includes more noise and natural effects such as illumination. Similar to MNIST, this dataset is labeled with respect to the digit category. In Figure \ref{fig:svhn_1}, latent codes with various interpretation are presented. In this experiment different amount of supervision result in different unsupervised representations retrieved. 

The SVHN dataset is perturbed by various noise factors such as blur and ambiguity in the digit categories. Figure \ref{fig:svhn_2} compares real samples with randomly generated synthetic samples by varying digit categories. The InfoGAN configuration ($0\%$ supervision) fails to encode a categorical latent code for the digit category. Leveraging some labeled information, our model becomes more robust to perturbations in the images. Through the introduction of labeled samples we are capable of exerting control over the latent space, encoding the digit labels in the categorical latent code $c_{ss}$. The smallest fraction of labeled data needed to achieve a notable effect is $10\%$ (i.e. $7'326$ labels out of $73'257$ samples). 

In Figure \ref{fig:svhn_3} we assess the performance of ss-InfoGAN with respect to $C$. The unsupervised configuration is left out since it is not able to control digit categories. As ss-InfoGAN exploits more label information, the generator converges faster and synthesizes more accurate images in terms of digit recognizability.
%%%%%%%%%%%%%% CELEBA
\begin{figure}[!htb]
	% Equal length
  \hspace*{\fill}%
  \subcaptionbox{$100\%$, Gender}{\includegraphics[width=4cm]{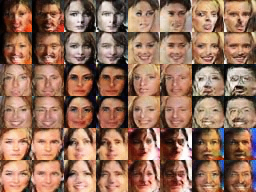}}\hfill%
  \subcaptionbox{$100\%$, Open mouth}{\includegraphics[width=4cm]{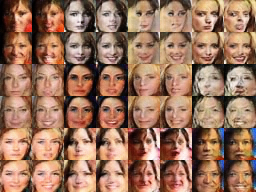}}\hfill%
  \subcaptionbox{$100\%$, Attractive}{\includegraphics[width=4cm]{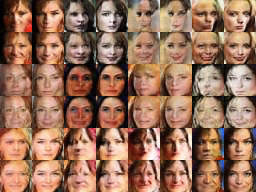}}\hfill
  \subcaptionbox{$1\%$, Gender}{\includegraphics[width=4cm]{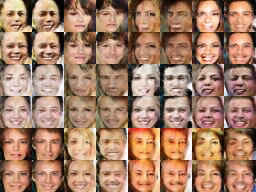}}\hfill%
  \subcaptionbox{$1\%$, Open mouth}{\includegraphics[width=4cm]{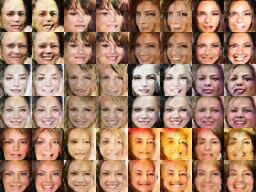}}\hfill%
  \subcaptionbox{$1\%$, Attractive}{\includegraphics[width=4cm]{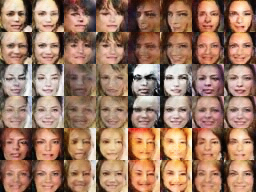}}%
  \hspace*{\fill}%
\caption{\textbf{Manipulating latent code on CelebA:} Synthetic samples generated by varying semi-supervised latent codes for the binary attributes "gender", "open mouth" and "attractive". Each $2\times2$ block corresponds to synthetic samples generated by keeping the input vector $(c_{ss},c_{us},z)$ fixed except the first semi-supervised categorical code encoding "smile" attribute (varied across the $y$-axis) and the other semi-supervised categorical latent code, whose interpretation is given in the caption (varied across the $x$-axis).}
	\label{fig:celeba_1}
\end{figure}
\begin{figure}[!htb]
  % Equal length
  \hspace*{\fill}%
  \subcaptionbox{Real samples}{\includegraphics[height=4.6cm]{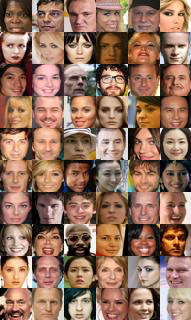}}\hfill%
  \subcaptionbox{$0\%$}{\includegraphics[height=4.6cm]{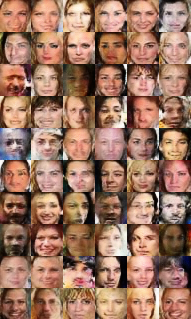}}\hfill%
  \subcaptionbox{$1\%$}{\includegraphics[height=4.6cm]{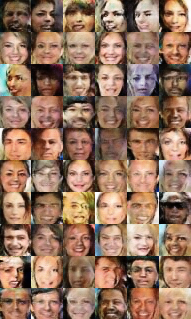}}\hfill%
  \subcaptionbox{$100\%$}{\includegraphics[height=4.6cm]{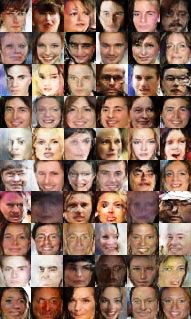}}%
  \hspace*{\fill}%
  \caption{\textbf{Random synthetic CelebA samples:} For each synthesized image the latent and noise variables are randomly drawn from their respective distribution.}
	\label{fig:celeba_2}
\end{figure}
\subsection{CelebA}
The CelebA dataset contains a rich variety of binary labels. We pre-process the data by extracting the faces via a face detector and then resize the extracted faces to $32\times32$. From the set of binary labels provided in the data we select the following attributes: "presence of smile", "mouth open", "attractive" and "gender".

Figure \ref{fig:celeba_1} shows synthetic images generated with ss-InfoGAN by varying certain latent codes. Although we experiment by using various hyper-parameters, InfoGAN is not able to learn an equivalent representation to these attributes. We see that for as low as $1\%$, $c_{ss}$ acquires the semantic meaning of $y$. This corresponds to $1'511$ labeled samples out of $151'162$. Figure \ref{fig:celeba_2} presents a batch of real samples from the dataset alongside with randomly synthesized samples from generators trained on various labeled percentages, with $0\%$ corresponding again to InfoGAN. 
 
\begin{figure}[!htb]
  % Equal length
  \hspace*{\fill}%
  \subcaptionbox{CelebA\label{fig:celeba_3}}{\includegraphics[width=6cm]{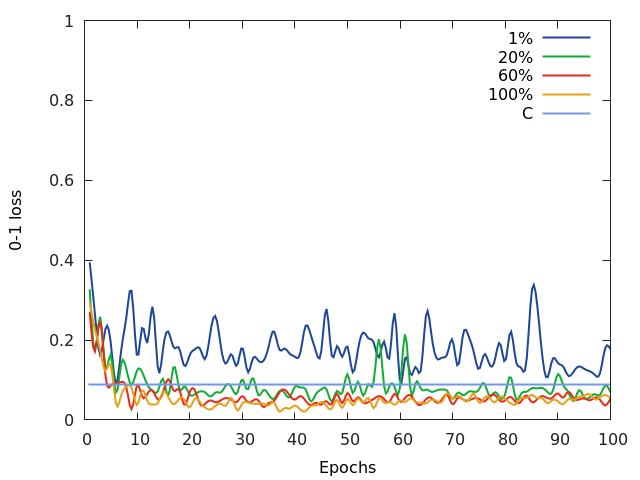}}\hfill%
  \subcaptionbox{CIFAR-10\label{fig:cifar10_3}}{\includegraphics[width=6cm]{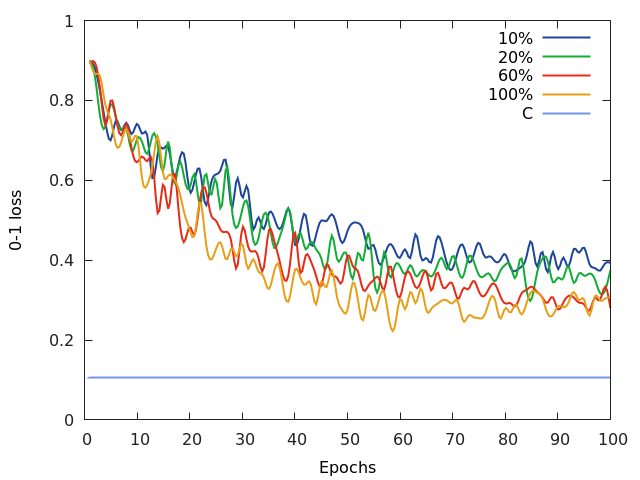}}%
  \hspace*{\fill}%
  \caption{\textbf{0-1 loss on synthetic samples}: (\ref{fig:celeba_3}) The model trained on CelebA dataset by leveraging the minimum amount of supervision that is sufficient to encode label ($1\%$) information shows unstable behavior. (\ref{fig:cifar10_3}) None of the models trained on CIFAR-10 dataset achieve the quality enough to reach real sample classification accuracy.}
\end{figure}

The performance of ss-InfoGAN on the independent classifier $C$ is shown in Figure \ref{fig:celeba_3}. For the lowest amount of labeling some instability can be observed. We believe this is due to the differences between the positives and the negatives of each binary label being more subtle than in other datasets. In addition, synthetic data generation exhibits certain variability which can obfuscate important parts of the image. However, using $20\%$ of labeled samples ensures a stable training performance.

%%%%%%%%%%%%%%%%%%%%%% CIFAR10
\subsection{CIFAR-10}
Finally we evaluate our model on CIFAR-10 dataset consisting of natural images. The data is labeled with the object category, which we use for the first semi-supervised categorical code. In order to stabilize training we apply instance noise \cite{sonderby2016}. 

\begin{figure}[!htb]
\centering
	  \includegraphics[width=0.8\linewidth]{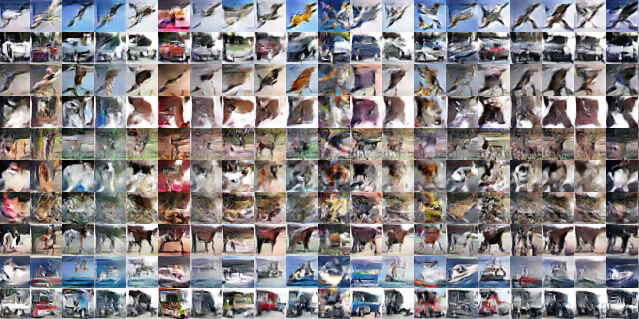}
	  \caption{\textbf{Manipulating latent code on CIFAR:} An example of varying an unsupervised latent code ($x$-axis) on CIFAR-10. Each row corresponds to a fixed code and represents class labels. The unsupervised latent code is not clearly interpretable.}
	  \label{fig:cifar10_1}
\end{figure}

On this dataset the unsupervised latent codes are not interpretable. An example is presented in Figure \ref{fig:cifar10_1} where the synthetic samples are generated by varying one of the unsupervised latent codes. Despite the fact that ss-InfoGAN model is trained by using \textit{all} label information, the semantic meaning of this unsupervised representation is not clear. The randomness of the natural images prevent models from learning interpretable representations in the absence of guidance.

\begin{figure}[!htb]
  % Equal length
  \hspace*{\fill}%
  \subcaptionbox{Real samples}{\includegraphics[height=4.6cm]{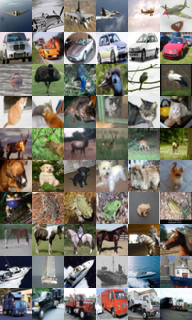}}\hfill%
  \subcaptionbox{$0\%$ \label{fig:cifar_samples_labelled_0_0}}{\includegraphics[height=4.6cm]{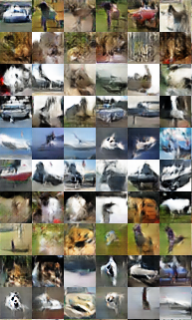}}\hfill%
  \subcaptionbox{$10\%$ \label{fig:cifar_samples_labelled_0_9}}{\includegraphics[height=4.6cm]{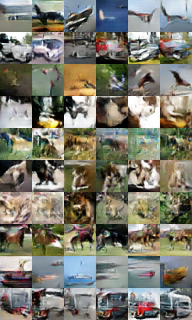}}\hfill%
  \subcaptionbox{$100\%$}{\includegraphics[height=4.6cm]{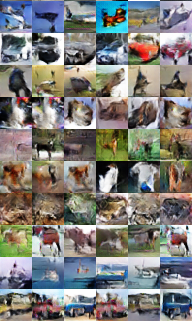}}%
  \hspace*{\fill}%
  \caption{\textbf{Random synthetic CIFAR-10 samples} Real samples and synthetic samples generated by the models trained with different amount of supervision. In each row, the semi-supervised categorical code encoding the image categories is kept fixed while rest of the input vector (i.e. $z$, $c_{us}$) and the remaining codes in $c_{ss}$, is randomly drawn from the latent distribution.}
	\label{fig:cifar10_2}
\end{figure}

Figure \ref{fig:cifar10_2} shows synthetic samples generated by models with different supervision configurations. InfoGAN has difficulties in learning the object category (see Figure \ref{fig:cifar_samples_labelled_0_0}) and hence in generating class-conditional synthetic images. For this dataset we find that labeling $10\%$ of the training data (corresponding to $5'000$ images out of $50'000$) is sufficient for ss-InfoGAN to encode class category (see Figure \ref{fig:cifar_samples_labelled_0_9}). 

In Figure \ref{fig:cifar10_3} classification accuracy of $C$ on the synthetic samples is plotted, again displaying the similar behavior of having better performance as more labels are available. It is evident that the additional information provided by the labels is fundamental to control \emph{what} the image depicts. We argue that attaining such low amounts of labels is feasible even for large and complicated datasets.

\subsection{Convergence speed of sample quality}
\begin{figure}[!htb]
	\begin{subfigure}{.4\textwidth}
		\centering
			\includegraphics[width=0.9\linewidth]{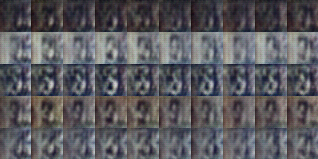}
		\caption{InfoGAN, epoch 26}
	\end{subfigure}%
	\begin{subfigure}{.4\textwidth}
		\centering
			\includegraphics[width=0.9\linewidth]{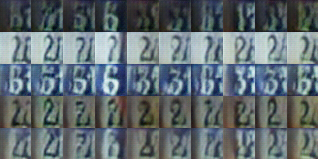}
		\caption{InfoGAN, epoch 47}
	\end{subfigure}
	\begin{subfigure}{.4\textwidth}
		\centering
			\includegraphics[width=0.9\linewidth]{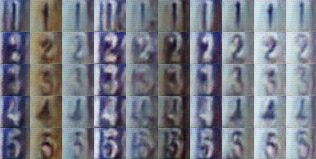}
		\caption{ss-InfoGAN, epoch 26}
	\end{subfigure}%
	\begin{subfigure}{.4\textwidth}
		\centering
			\includegraphics[width=0.9\linewidth]{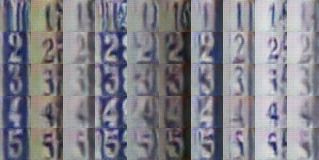}
		\caption{ss-InfoGAN, epoch 47}
	\end{subfigure}
	\caption{Samples from InfoGAN and ss-InfoGAN trained on SVHN at two different epochs}
	\label{fig:conv_sample_quality}
\end{figure}

During the course of the experiments, it is observed that the convergence of synthetic sample quality is faster in comparison to InfoGAN. Figure \ref{fig:conv_sample_quality} shows synthetic SVHN samples from a fully supervised ss-infoGAN and infoGAN at training epoch $26$ and $47$. The training epochs are chosen by inspection so that each model starts producing recognizable images. Therefore we can quantitatively say that ss-InfoGAN converges faster than InfoGAN.

%% file: sec/conclusion.tex
% !TEX root = ../template.tex

\section{Conclusion}
We have introduced ss-InfoGAN a novel semi-supervised generative model. We have shown that including few labels increases the convergence speed of the latent codes $c_{ss}$ and that these represent the same meaning as the labels $y$. This speed-up increases as more data samples are labeled. Although in theory this only improves convergence speed of $c_{ss}$, we have shown empirically that the sample quality convergence speed has improved as well.

In addition, it was shown that using labeling information is useful in cases where InfoGAN fails to find a \emph{specific} representation, such as in the case of SVHN, CelebA and CIFAR-10. To successfully guide a latent code to the desired representation, it is sufficient that the dataset contains only a minimal subset of labeled data. The amount of required labels ranges from $0.22\%$ for the simplest datasets (MNIST) to a maximum of $10\%$ for the most complex datasets (CIFAR-10). We argue that acquiring such low percentages of labels is cost effective and makes the proposed architecture an attractive choice if control over \emph{specific} latent codes is required and full supervision is not an option.

\subsubsection{Acknowledgements}
This work was supported in parts by the ERC grant OPTINT (StG-2016-717054)